\documentclass{svmult}
\usepackage{mathptmx}       
\usepackage{helvet}         
\usepackage{courier}        
\usepackage{type1cm}        
\usepackage{graphicx}        
\usepackage{multicol}        
\usepackage[bottom]{footmisc}
\usepackage{amsmath,amssymb,epsfig}
\usepackage{wrapfig}

\usepackage{color}
\usepackage{psfrag}

\newcommand{\assign}{:=}
\newcommand{\tmem}[1]{{\em #1\/}}
\newcommand{\tmop}[1]{\ensuremath{\operatorname{#1}}}
\newcommand{\tmtextbf}[1]{{\bfseries{#1}}}
\newenvironment{itemizedot}{\begin{itemize} \renewcommand{\labelitemiv}{$\bullet$}}{\end{itemize}}
\definecolor{grey}{rgb}{0.75,0.75,0.75}
\definecolor{orange}{rgb}{1.0,0.5,0.5}
\definecolor{brown}{rgb}{0.5,0.25,0.0}
\definecolor{pink}{rgb}{1.0,0.5,0.5}

\let\sect\section
\let\subsect\subsection
\renewcommand\section[1]{\vspace*{-0.5cm}\sect{#1} \vspace*{-0.5cm}}
\renewcommand\subsection[1]{\vspace*{-0.5cm}\subsect{#1} \vspace*{-0.5cm}}

\begin{document}
\title*{Cusp points in the parameter space of R\underline{P}R-2P\underline{R}R parallel manipulators}
\author{G. Moroz$^{1,2}$, D. Chablat$^1$, P. Wenger$^1$, F. Rouiller$^2$}
\institute{$^1$Institut de Rercheche en Communications et Cybern\'etique de Nantes, France, \email{(Guillaume.Moroz, Damien.Chablat, Philippe.Wenger)@irccyn.ec-nantes.fr} \\
$^2$Laboratoire d'informatique de Paris, France, \email{Fabrice.Rouiller@inria.fr}}
\maketitle

\abstract{This paper investigates the existence conditions of cusp points in the design parameter space of the R\underline{P}R-2P\underline{R}R parallel manipulators. Cusp points make possible non-singular assembly-mode changing motion, which can possibly increase the size of the aspect, i.e. the maximum singularity free workspace. The method used is based on the notion of discriminant varieties and Cylindrical Algebraic Decomposition, and resorts to  Gr\"obner bases for the solutions of systems of equations.}


\section{Introduction}
It is well known that the workspace of a parallel manipulator is divided into singularity-free connected regions \cite{chablat1998working}. These regions are separated by the so-called parallel singular configurations, where the manipulator loses its stiffness and gets out of control. The so-called cuspidal manipulators have the ability to change their assembly-mode without running into a singularity, which thus may increase the size of the singularity-free regions \cite{McaD99, Hck09}. The word ``cuspidal'' stems from the notion of cuspidal configuration, defined as one configuration where three direct kinematic solutions coalesce. 
A cuspidal configuration in the manipulator joint space allows non-singular assembly-mode changing motions. Thus, determining cuspidal configurations is an important issue that has attracted the attention of several researchers \cite{McaD99,HAPMieee09,ZWCr07,BWSieee08}. In particular, \cite{ZWCr07} (resp. \cite{BWSieee08}) has analyzed the cuspidal configurations of planar 3-R\underline{P}R (resp. 3-\underline{P}RR) manipulators\footnote{The underlined letter means an actuated joint}. More recently, \cite{HAPMieee09} studied the R\underline{P}R-2P\underline{R}R, a simpler planar 3-DOF manipulator that lends itself to algebraic calculus \cite{HAPMieee09}. In both papers, the cusp configurations were determined by looking for the triple roots of a univariate polynomial. This approach may yield spurious solutions. In this paper, the cuspidal configuration are determined directly from the Jacobian of the whole set of geometric constraints of the robot, which guaranties that only true solutions are obtained. Then, we classify the parameter space of a family of R\underline{P}R-2P\underline{R}R manipulators according to the number of cuspidal configurations. It is shown that these manipulators have either  $0$ or $16$ cuspidal configurations. The proposed method is based on the notion of discriminant varieties and cylindrical algebraic decomposition, and resorts to Gr\"obner bases for the solutions of systems of equations. 
%
%
\section{Mechanism under study}
\vspace*{0.5cm}
\subsection{Kinematic equations}
\begin{wrapfigure}{r}{55mm}
\vspace*{-1cm}
  \begin{center}
     \psfrag{A1}{$A_1$}
     \psfrag{A2}{$A_2$}
     \psfrag{A3}{$A_3$}
     \psfrag{B1}{$B_1(x,y)$}
     \psfrag{B2}{$B_2$}
     \psfrag{B3}{$B_3$}
     \psfrag{T1}{$\theta_1$}
     \psfrag{T2}{$\theta_2$}
     \psfrag{T3}{$\theta_3$}
     \psfrag{T} {$\alpha$}
     \psfrag{P1}{$\rho_1$}
     \psfrag{P2}{$\rho_2$}
     \psfrag{P3}{$\rho_3$}
     \psfrag{a}{a}
     \psfrag{b}{b}
     \psfrag{L2}{$L_2$}
     \psfrag{L3}{$L_3$}
     \includegraphics[width=4cm]{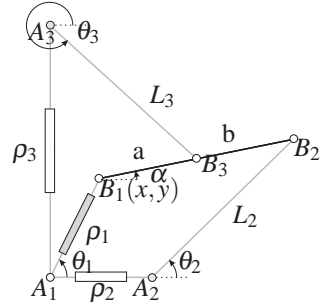}
    \caption{A R\underline{P}R-2P\underline{R}R parallel manipulators with ($a=1$, $b=2$, $L_2=2$, $L_3=2$, $x=1/2$,$y=1$, $\theta=0.2$). The actuated joint symbols were filled in gray.}
    \label{Figure1}
  \end{center}
\vspace*{-1cm}
\end{wrapfigure}
A R\underline{P}R-2P\underline{R}R parallel manipulator is shown in Fig.~\ref{Figure1}. This manipulator was analyzed in \cite{HAPMieee09}. It has 1 actuated prismatic joint $\rho_1$, and 2 passive prismatic joints $\rho_2$ and $\rho_3$. The two revolute joints centered in $A_2$ and $A_3$ are actuated while the ones centered in $A_1$, $B_1$, $B_2$ and $B_3$ are passive. The pose of the moving platform is described by the position coordinates $(x, y )$ of $B_1$ and by the orientation $\alpha$ of the moving platform $B_1B_2$. The input variables (actuated joints values) are defined by $\rho_1$, $\theta_2$ and $\theta_3$. The points $B_1$, $B_2$ and $B_3$ are aligned, a= ($B_1$, $B_2$), b= ($B_1$, $B_3$), $L_2= (A_2, B_2)$ and $L_3= (A_3, B_3)$.

The geometric constraints can be expressed by the following 5 equations \cite{HAPMieee09}:
{\footnotesize
\begin{align}
f_1: \rho_1^2 =&x^2+y^2 \notag \\
f_2: x =& \rho2 + L_2 \cos(\theta_2) - b \cos(\alpha) &
f_4: x =& L_3 \cos(\theta_3) - a \cos(\alpha) \\
f_3: y =& L_2 \sin(\theta_2) - b \sin(\alpha) &
f_5: y =& \rho3 + L_3 \sin(\theta_3) - a \sin(\alpha) \notag
\end{align} }
Without loss of generality, we fix $a=1$ in the rest of the article.
%
%

\subsection{An algebraic model}

The singular and cuspidal equations were previously computed using the three
following steps \cite{HAPMieee09}:
\begin{enumerate}
  \item Reduce the equation system to a polynomial equation depending on the 
  articular variables $\rho_1,\theta_1,\theta_2$ and one pose variable $\alpha$:
  $g (\rho_1, \theta_1, \theta_2, \alpha) = 0$ (this step is done
  by eliminating variables, either with resultants \ or with Gr\"obner basis )
  
  \item Add the constraint $\frac{\partial g}{\partial \alpha} = 0$ to define
  the parallel singularities
  
  \item Add the constraint $\frac{\partial g}{\partial \alpha} = 0$ and
  $\frac{\partial^2 g}{\partial \alpha^2} = 0$ to define the cuspidal
  configurations
\end{enumerate}
This approach has the advantage of reducing the problem of computing the cusp
configurations, to the problem of analysing the triple roots of a single polynomial.
However, this only gives a necessary condition for the manipulator to have
cusp configurations. In particular it is possible that $3$ configurations of the robot coalesce in one coordinate but not in the others.

Let us come back to the theoretical definitions, using Jacobian matrices to
define directly the triple roots of the original system of equations in all
the input and output variables.  If $P$ is a list of polynomials and $X$ a list of variables, let $J_k (P, X)$ be the union of $P =\{p_1, \ldots, p_m \}$ and of all the $k \times k$ minors of the Jacobian matrix of the $p_i$ with respect to the $X_i$.

For the analysis of the RPR--2PRR manipulator, we introduce:

$Y := [x, y, \alpha, \rho_2, \rho_3] \quad\quad 
 W \assign [b, L_2, L_3, \rho_1, \theta_2, \theta_3] \quad\quad
 S \assign \{f_1, \ldots, f_5 \}$.

Using these notations, the parallel singularities of the manipulator are
defined by $\{\mathbf{v} \in \mathbb{R}^{11}, p(\mathbf{v}) = 0, q(\mathbf{v}) > 0, \forall p \in J_5 (S, Y),
\forall q \in \{b,L_2,L_3,\rho_1\}$ so that the cuspidal configurations are fully characterized
by :

\begin{center}
  $\mathcal{S} =\{\mathbf{v} \in \mathbb{R}^{11}, p(\mathbf{v}) = 0, q(\mathbf{v}) > 0, \forall p \in J_5 (J_5
  (S, Y), Y), \forall q \in \{b,L_2,L_3,\rho_1\}\}$
\end{center}

\section{Main tools from computational algebra}

The algebraic problem to be solved is
basically related to the resolution of polynomial parametric systems.

More specifically, one needs to solve a system of the following form :

\begin{center}
  $E =\{\mathbf{v} \in \mathbb{R}^n, p_1 (\mathbf{v}) = 0, \ldots, p_m (\mathbf{v}) = 0, q_1
  (\mathbf{v}) > 0, \ldots q_l (\mathbf{v}) > 0\}$
\end{center}

where $p_1, \ldots, p_m, q_1, \ldots, q_l$ are polynomials with rational
coefficients depending on the unknowns $X = [X_1, \ldots, X_n]$ and on the
parameters $U = [U_1, \ldots, U_d]$.

There are numerous possible ways of solving parametric
systems in general. Here we focus on the use of Discriminant Varieties (DV, \cite{LR07}) and
Cylindrical Algebraic Decomposition (CAD, \cite{Cbook75}) for two reasons.  It provides a formal decomposition of the parameter space through an exactly known algebraic variety (no approximation).  It has been already successfully used for similar mathematical classes of problems (see {\cite{CR02}}).

To reduce the dimension of the parameter space to three so that it can be displayed, we set $L_2 = L_3$. Not that the proposed method can treat the general case $L_2\neq L_3$ without any problems.
When $L_2 = L_3$, the system to solve is $\mathcal{S}$ with the unknowns $[x, y, \alpha, \rho_2, \rho_3, \theta_2, \theta_3]$ and the parameters $[b, L_2, \rho_1]$.

\subsection{Basic black-boxes}

First experiments are often performed for specific values of the parameters,
especially singular and/or degenerated cases. Here, we mainly use exact
computations, namely formal elimination of variables (resultants, Gr\"obner bases) and resolution of systems with a finite number of solutions, including
  univariate polynomials.

Let us describe the global solver for zero-dimensional
systems. It will be used as a black box in the general algorithm we describe
in the sequel.

Given any system of equations $p_1 = 0, \ldots, p_m = 0$ of polynomials of
$\mathbb{Q}[X_1, \ldots, X_n]$, one first computes a Gr\"obner basis of the
ideal $< p_1, \ldots, p_m >$ for any ordering.

At this stage, one can detect easily if the system has or has not finitely many
complex solutions.

If yes, then compute a so called Rational Univariate Representation or RUR (see
{\cite{Raaecc99}}) of $< p_1, \ldots, p_m >$, which is, in short, an equivalent
system of the form : $\{f (T) = 0, X_1 = \frac{g_1 (T)}{g (T)}, \ldots, X_n =
\frac{g_n (T)}{g (T)} \}$, $T$ being a new variable that is independent of
$X_1, \ldots, X_n$, equipped with a so called {\tmem{separating element}}
(injective on the solutions of the system) $u \in \mathbb{Q}[X_1, \ldots,
X_n]$ and such that :
$$\begin{array}{ccccc}
  V (\left< p_1, \ldots, p_m \right>) & \xrightarrow{u} & V (f) &
  \xrightarrow{u^{-1}} & V(\left< p_1,\ldots,p_m\right>)\\
  (x_1, \ldots, x_n) & \mapsto & \beta=u (x_1, \ldots, x_n) &
  \mapsto & \left( \frac{g_1 (\beta)}{g (\beta)}, \ldots, \frac{g_n (\beta)}{g (\beta)}  \right)
\end{array}$$
defines a bijection between the (real) roots of the system
(denoted by $V (p_1, \ldots, p_m)$) and the (real) roots of the univariate
polynomial (denoted by $V (f)$).

We then solve the univariate polynomial $f$, computing so called isolating
intervals for its real roots, say non-overlapping intervals with rational
bounds that contain a unique real root of $f$ (see {\cite{RZ03}}). Finally,
interval arithmetic is used for getting isolating boxes of the real roots of the
system (say non overlapping products of intervals with rational bounds
containing a unique real root of the system), by studying the RUR over the
isolating intervals of $f$. In practice, we use the function {\tmem{RootFinding[Isolate]}} from Maple software, which performs exactly the computations described above.

\begin{figure}
\vspace*{-0.5cm}
\begin{center}
\psfrag{A1}{$A_1$}
\psfrag{A3}{$A_3$}
\psfrag{A2}{$A_2$}
\psfrag{B1}{$B_1$}
\psfrag{B2}{$B_2$}
\psfrag{B3}{$B_3$}
\includegraphics[width=0.2\textwidth]{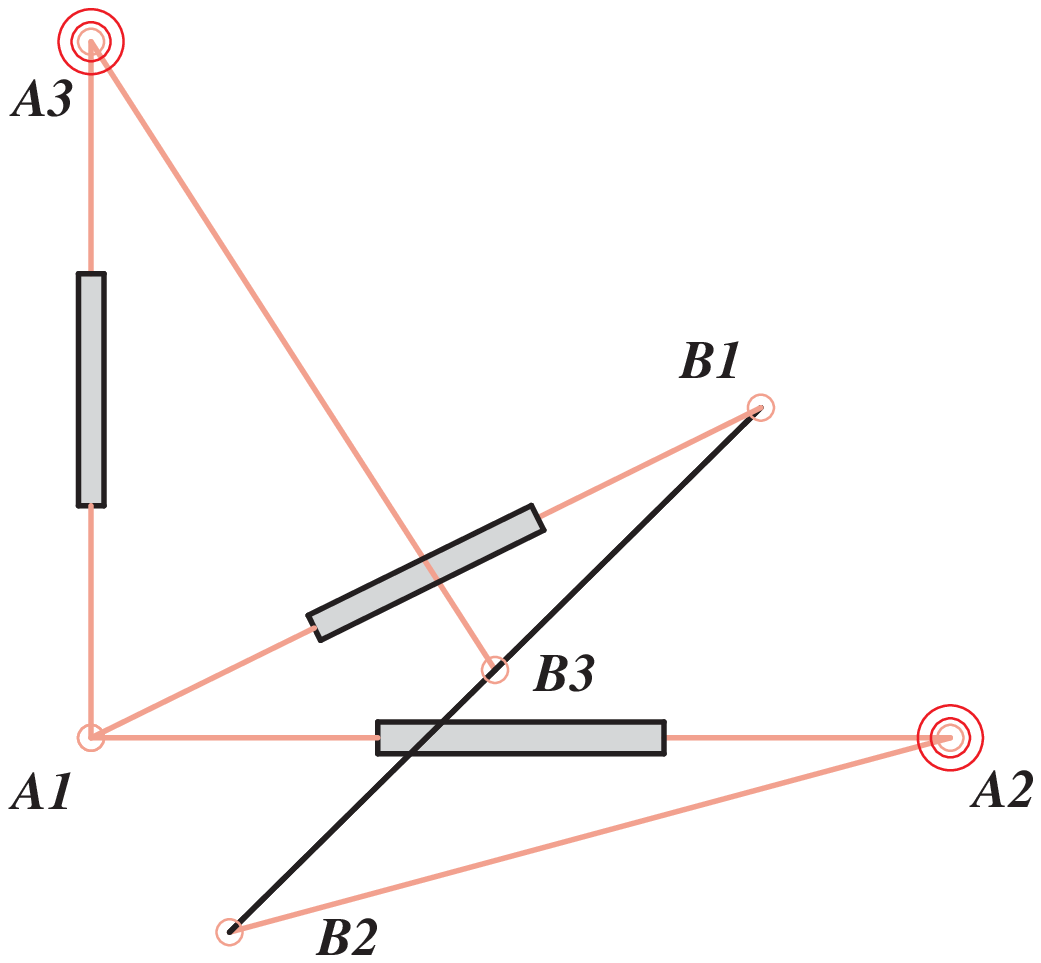}
\hspace{0.2\textwidth}
\includegraphics[width=0.2\textwidth]{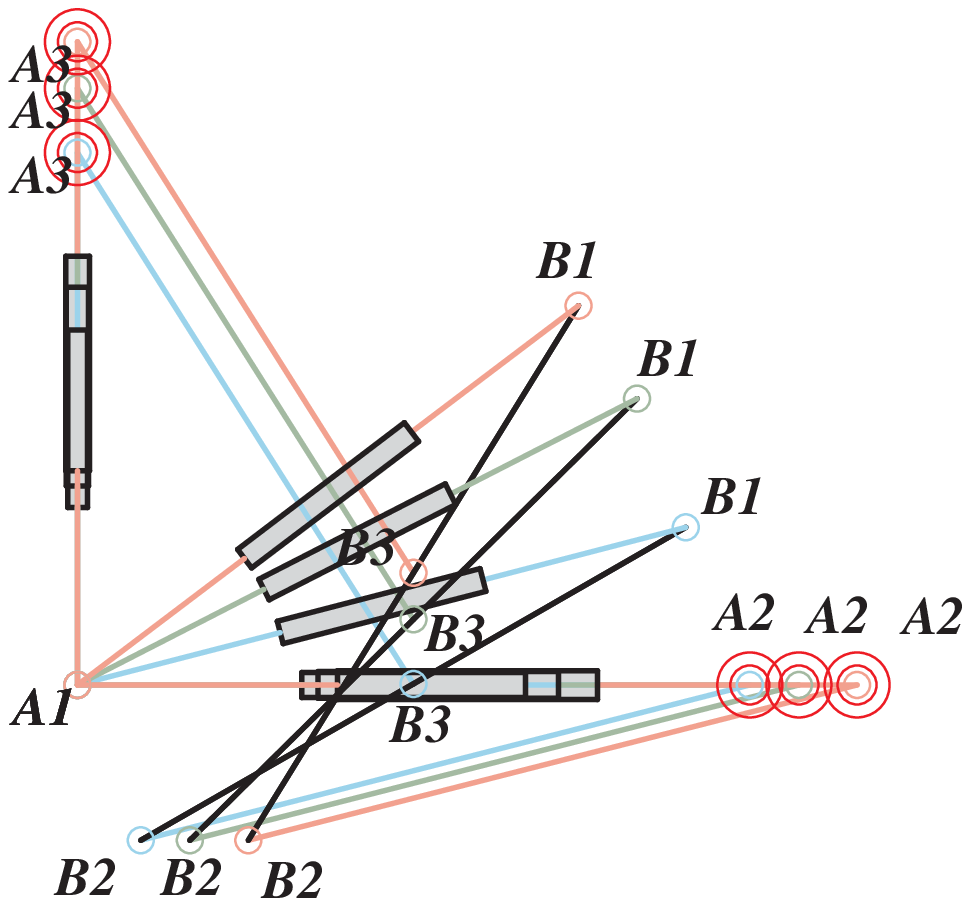}
\caption{A cuspidal configuration (left), and the three converging configurations (right).}
\label{figure:cuspconf}
\end{center}
\vspace*{-1cm}
\end{figure}

For example, with ($b=2$, $L_2=2$, $L_3=3$, $\rho_1=2$), the polynomial system $\mathcal{S}$ defining the cuspidal configurations has 16 real solutions. 
One of these solutions is $(\rho_2=2.30,\rho_3=1.86,\phi=-2.36, x=1.79, y=0.885,\theta_2=-2.88,\theta_3=-0.999)$.  We observe the three coalescing configurations around this root by calculating the direct kinematic solutions with $\theta_2=-2.892$ and $\theta_3=-1.007$. These solutions are shown in Figure \ref{figure:cuspconf}.

\subsection{\tmtextbf{Discriminant varieties}}

The above method allows one to study instances of the problem and may be used
together with a discretization of the parameter space to get a first idea of
the complexity of the general problem to be solved. But the true issue
addressed in this paper is to find criteria on the parameters that allow
classifying the configurations to be studied (for example to distinguish the
manipulators having cuspidal configurations from the others). This leads to a more
general problem since one then has to study non zero-dimensional,
semi-algebraic sets.

Let $p_1, \ldots, p_m, q_1, \ldots, q_l$ be polynomials with rational
coefficients depending on the unknowns $X_1, \ldots, X_n$ and on the
parameters $U_1, \ldots, U_d$. Let us consider the constructible set :

\begin{center}
  $\mathcal{C} =\{\mathbf{v} \in \mathbb{C}^n \hspace{0.75em}, \hspace{0.75em} p_1 (\mathbf{v}) = 0, \ldots, p_m (\mathbf{v}) = 0, q_1 (\mathbf{v}) \neq 0, \ldots q_l (\mathbf{v}) \neq 0\}$
\end{center}

If we assume that $\mathcal{C}$ is a finite number of points for almost all
the parameter values, a discriminant variety $V_D$ of $\mathcal{C}$ is a
variety in the parameter space $\mathbb{C}^d$ such that, over each connected
open set $\mathcal{U}$ satisfying $\mathcal{U} \cap V_D = \emptyset$,
$\mathcal{C}$ defines an analytic covering. In particular, the number of
points of $\mathcal{C}$ over any point of $\mathcal{U}$ is constant.

Let us now consider the following semi-algebraic set :

\begin{center}
  $\mathcal{S} =\{\mathbf{v} \in \mathbb{R}^{11}, p(\mathbf{v}) = 0, q(\mathbf{v}) > 0, \forall p(\mathbf{v}) \in J_5 (J_5
  (S, Y), Y), \forall q(\mathbf{v}) \in \{b,L_2,L_3,\rho_1\}\}$
\end{center}

If we assume that $\mathcal{S}$ has a finite number of solutions over at least
one real point that does not belong to $V_D$, then $V_D \cap \mathbb{R}^d$ can
be viewed as a real discriminant variety of $\mathcal{S}$, with the same
property : over each connected open set $\mathcal{U} \subset \mathbb{R}^d$
such that $\mathcal{U} \cap V_D = \emptyset$, $\mathcal{C}$ defines an
analytic covering. In particular, the number of points of
$\mathbb{R}$ over any point of $\mathcal{U}$ is constant.

Discriminant varieties can be computed using basic and well known tools from
computer algebra such as Gr\"obner bases (see {\cite{LR07}}) and a full
package computing such objects in a general framework is available in Maple
software through the {\tmem{RootFinding[Parametric]}} package. Figure \ref{figure:dv} represents the discriminant variety of the cuspidal configurations of the R\underline{P}R-2P\underline{R}R manipulator.



\subsection{\tmtextbf{The complementary of a discriminant variety}}
\label{section:CAD}
At this stage, we know, by construction, that over any simply connected open
set that does not intersect the discriminant variety (so-called regions), the system has a constant number of (real) roots.


The goal of this part is now to provide a description of the regions for which the number of solutions of the system at hand is constant. For that, we compute an open CAD (\cite{Cbook75,DSSissac04}).

Let $\mathcal{P}_d \subset \mathbb{Q}[U_1, \ldots, U_d]$ be a set of
polynomials. For $i = d-1 \ldots 0$, \ we introduce a set of polynomials
$\mathcal{P}_{i} \subset \mathbb{Q}[U_1, \ldots, U_{d - i}]$ defined by a backward recursion:
\begin{itemizedot}
  \item $\mathcal{P}_d$ : the polynomials defining the discriminant variety
  \item $\mathcal{P}_i$ : $\{\tmop{Discriminant} (p, U_i)$,$
  \tmop{LeadingCoefficient} (p, U_i)$, $\tmop{Resultant} (p, q, U_i)$,\\\indent$\quad\quad p, q \in
  \mathcal{P}_{_{i + 1}} \}$
\end{itemizedot}

\begin{figure}[h]
\begin{minipage}[b]{0.45\linewidth}
\centering
\psfrag{S1}{$\rho_1$}
\psfrag{L2}{$L_2$}
\psfrag{b}{$b$}
\psfrag{0}[t][t][0.7]{0}
\psfrag{1}[t][t][0.7]{1}
\psfrag{2}[t][t][0.7]{2}
\psfrag{3}[t][t][0.7]{3}
\psfrag{4}[t][t][0.7]{4}
\psfrag{5}[t][t][0.7]{5}
\psfrag{6}[t][t][0.7]{6}
\psfrag{7}[t][t][0.7]{7}
\psfrag{8}[t][t][0.7]{8}
\psfrag{10}[t][t][0.7]{10}
\includegraphics[width=\textwidth]{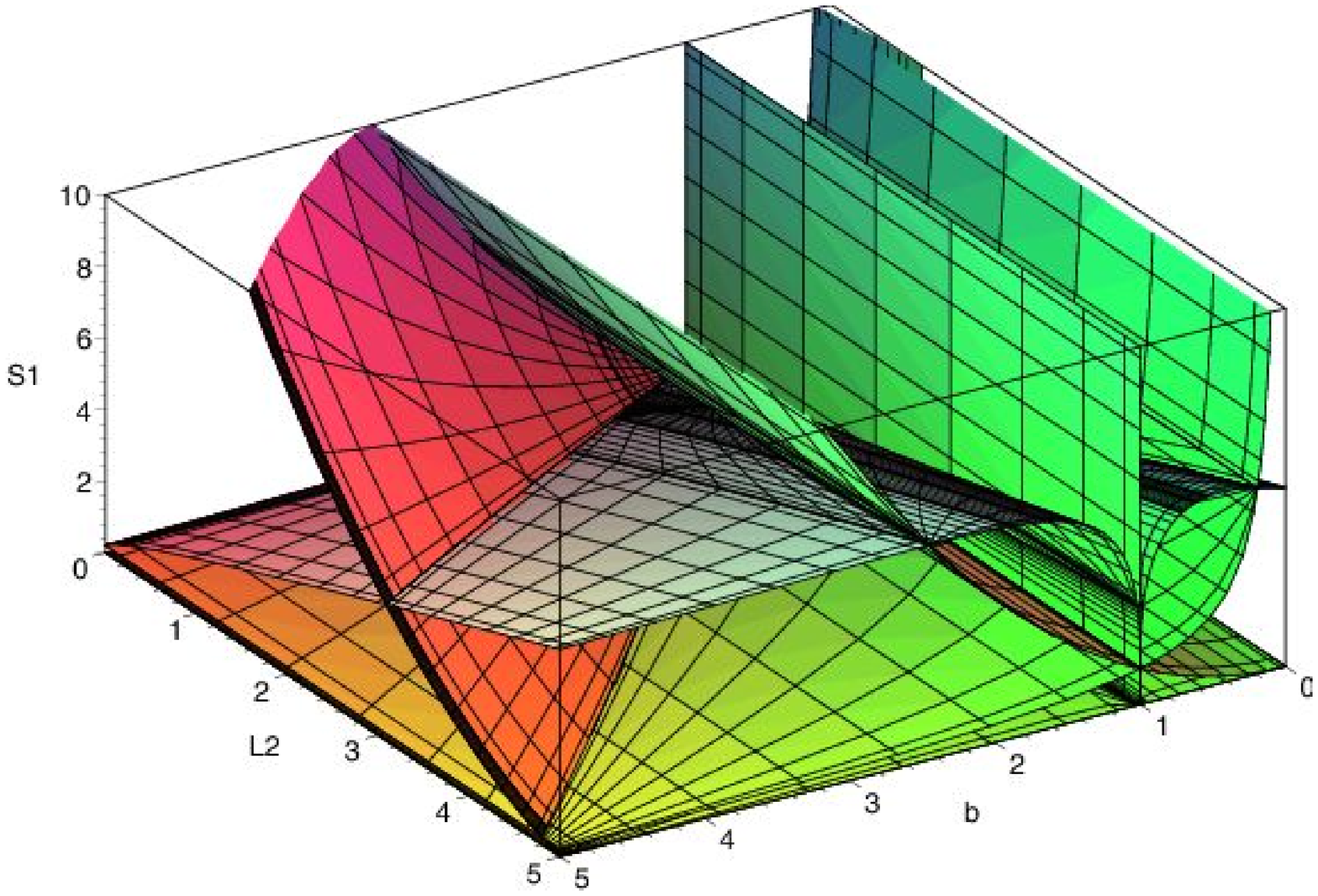}
\caption{Discriminant Variety of the cuspidal configurations}
\label{figure:dv}
\end{minipage}
\hspace{0.09\linewidth}
\begin{minipage}[b]{0.45\linewidth}
\centering
\psfrag{0}[t][t][0.7]{0}
\psfrag{1}[t][t][0.7]{1}
\psfrag{2}[t][t][0.7]{2}
\psfrag{3}[t][t][0.7]{3}
\psfrag{4}[t][t][0.7]{4}
\psfrag{5}[t][t][0.7]{5}
\psfrag{6}[t][t][0.7]{6}
\psfrag{7}[t][t][0.7]{7}
\psfrag{8}[t][t][0.7]{8}
\psfrag{10}[t][t][0.7]{10}
\psfrag{S1}{$\rho_1$}
\psfrag{L2}{$L_2$}
\psfrag{b}{$b$}
\includegraphics[width=\textwidth]{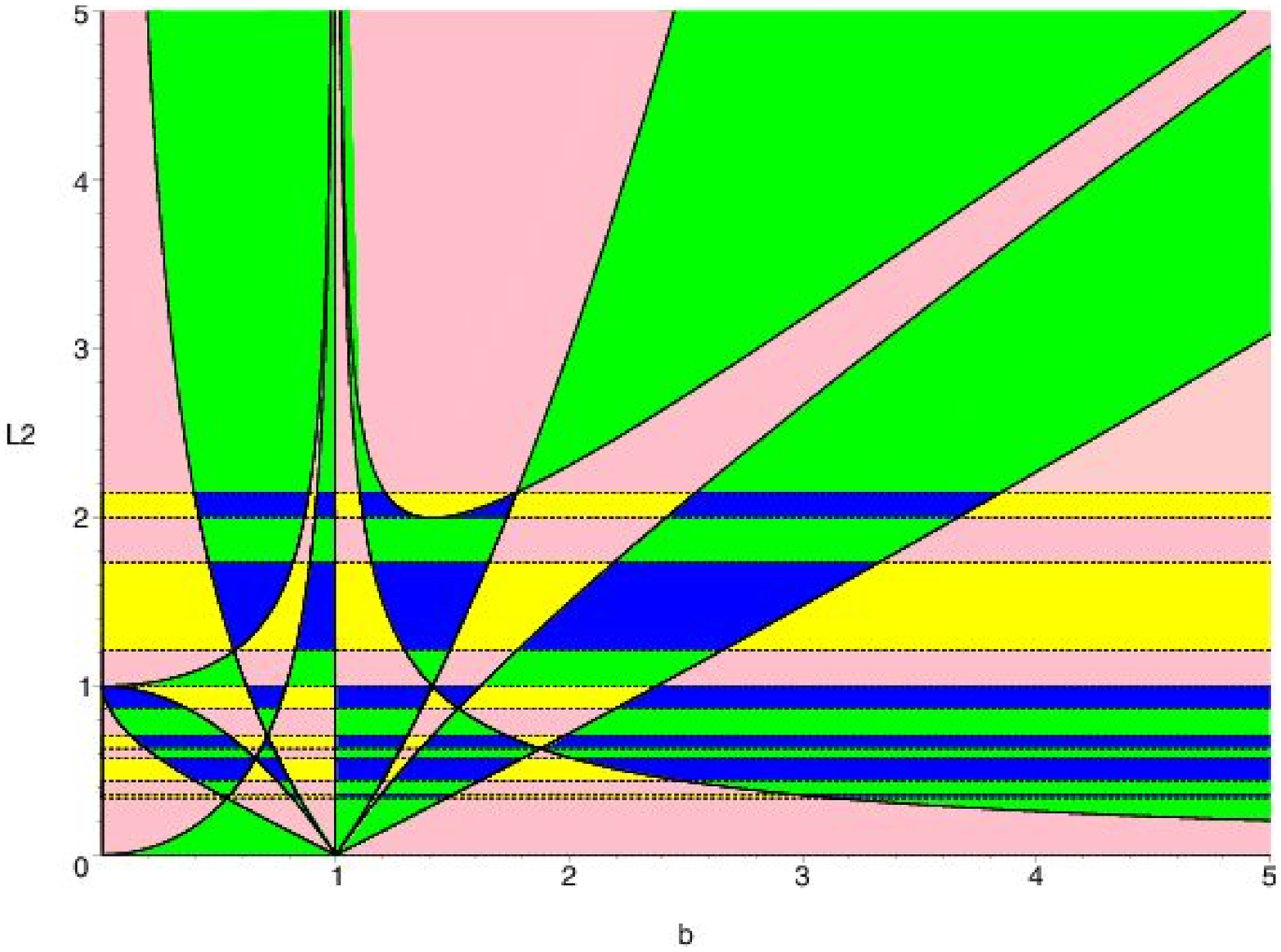}
\caption{$V_2$ for the cuspidal configurations of the manipulator.}
\label{figure:cad}
\end{minipage}

\vspace*{-0.7cm}
\end{figure}

We can associate to each $\mathcal{P}_i$ an algebraic variety of dimension at most $i - 1$ : $V_{i} = V ( \prod_{p \in \mathcal{P}_{i}} p)$. Figure \ref{figure:dv} and \ref{figure:cad} represent respectively $V_3$ and $V_2$ for the manipulator at hand.

The $V_i$ are used to define recursively a finite union of simply connected open subsets of $\mathbb{R}^{i}$ of dimension $i$: $\cup^{n_{i}}_{k = 1} \mathcal{U}_{i, k}$
  such that $V_{i} \cap \mathcal{U}_{i, k} = \emptyset$, and one point
  $u_{i, k}$ with rational coordinates in each $\mathcal{U}_{i, k}$.

In order to define the $\mathcal{U}_{i,k}$, we introduce the following notations. If $p$ is a univariate polynomial with $n$ real roots:

{\hfill $Root(p,l)=\left\{\begin{array}{l}
    -\infty\mbox{ if }l\leq 0\\
    \mbox{the $l^{th}$ real roots of $p$ if }1\leq l \leq n\\
    +\infty\mbox{ if }l>n \end{array}\right. $ \hfill}

Moreover, if $p$ is a $n$-variate polynomial, and $\mathbf{v}$ is a $n-1$-uplet, then $p^\mathbf{v}$ denotes the univariate polynomial where the first $n-1$ variables have been replaced by $\mathbf{v}$.

Roughly speaking, the recursive process defining the $\mathcal{U}_{i,k}$ is the following:
\begin{itemizedot}
 \item For $i=1$, let $p_1 = \prod_{p \in \mathcal{P}_1} p$. Taking $\mathcal{U}_{1, k} =] Root(p,k) ; Root(p,k+1) [$ for $k$ from $0$ to $n$ where $n$ is the number of real roots of $p_1$, one gets a partition of $\mathbb{R}$ that fits the above definition. Moreover, one can chose arbitrarily one rational point $u_{1, k}$ in each $\mathcal{U}_{1, k}$.
\item Then, let $p_i = \prod_{p \in \mathcal{P}_1} p$. The regions $\mathcal{U}_{i,k}$ and the points $u_{i,k}$ are of the form:
$$\begin{array}{ll@{}l}
\mathcal{U}_{i,k}&=\left\{(v_1,...,v_{i-1},v_i) \mid \right.&\mathbf{v}:=(v_1,...,v_{i-1})\in\mathcal{U}_{i-1,j},\\
&&\left. v_i\in ]Root(p_i^{\mathbf{v}},l), Root(p_i^{\mathbf{v}},l+1)[ \right\}\\
u_{i,k}&=(\beta_1,...,\beta_{i-1},\beta_i),&\mbox{ with }\left\{\begin{array}{l} 
    (\beta_1,...,\beta_{i-1})=u_{i-1,j}\\
    \beta_i\in]Root(p_i^{u_{i-1,j}},l),Root(p_i^{u_{i-1,j}},l+1)[ \end{array}\right.
\end{array}
$$
where $j,l$ are fixed integer.
\end{itemizedot}


For our example, we get for $p_3$ a trivariate polynomial of degree 33, for $p_2$ a bivariate polynomial of degree 113, and for $p_1$ a univariate polynomial of degree 59. The zero-dimensional solver then provides the positive real roots of $p_1$ (Table \ref{table:b}), from which we easily deduce the open intervals $u_{1, k}$.  We then use the zero-dimensional solver to solve every $p_2 (u_{1, k}, U_2)$ and deduce all the tests points of the $\mathcal{U}_{2, k'}$ in each cells of Figure \ref{figure:cad}. Finally we use the zero-dimensional solver to solve every $p_3 (u_{2, k}, U_3)$ and deduce all the tests points of the $\mathcal{U}_{3, k'}$, describing so the complementary of the discriminant variety.

\begin{table}
\vspace*{-0.5cm}
\begin{center}
\caption{Numerical values of the positive roots of $p_1$}
\begin{tabular}{|l||c|c|c|c|c|c|c|c|c|c|c|c|}
\hline
b & $b_1$ & $b_2$ & $b_3$ & $b_4$ & $b_5$ & $b_6$ & $b_7$ & $b_8$ & $b_9$ & $b_{10}$ & $b_{11}$ & $b_{12}$ \\
\hline
  & 0.0 & 0.533& 0.564& 0.617& 0.656& 0.707& 1.0& 1.41 & 1.52& 1.62& 1.77& 1.88 \\
\hline
\end{tabular}
\label{table:b}
\end{center}
\vspace*{-1cm}
\end{table}

\subsection{\tmtextbf{Discussing the number of solutions of the parametric
system}.}

At this stage, we have a full description of the complementary of the
discriminant variety of the system to be solved : a recursive process for the
construction of each cell $\mathcal{U}_{d, k}$ and a test point (with
rational coordinates) in each of these cells. By definition of the
discriminant variety, we know that the system has a constant finite number of
solutions over each of these cells and computing this
number for each cell is the only remaining step. This can be done simply by solving all the
systems $\mathcal{S}_{|U = u_{d, k}}$, $k = 1 \ldots n_d$ using the
zero-dimensional solver.

\begin{figure}
\vspace*{-0.5cm}
\begin{center}
\psfrag{S1}{\hspace{-1.1ex}$\rho_1$}
\psfrag{L2}{\hspace{-4ex}$L_2$}
\psfrag{b}{\hspace{2ex}$b$}
\psfrag{0}[t][t][0.5]{0}
\psfrag{1}[t][t][0.5]{1}
\psfrag{2}[t][t][0.5]{2}
\psfrag{3}[t][t][0.5]{3}
\psfrag{4}[t][t][0.5]{4}
\psfrag{5}[t][t][0.5]{5}
\psfrag{6}[t][t][0.5]{6}
\psfrag{7}[t][t][0.5]{7}
\psfrag{8}[t][t][0.5]{8}
\includegraphics[height=0.4\textwidth,angle=-90]{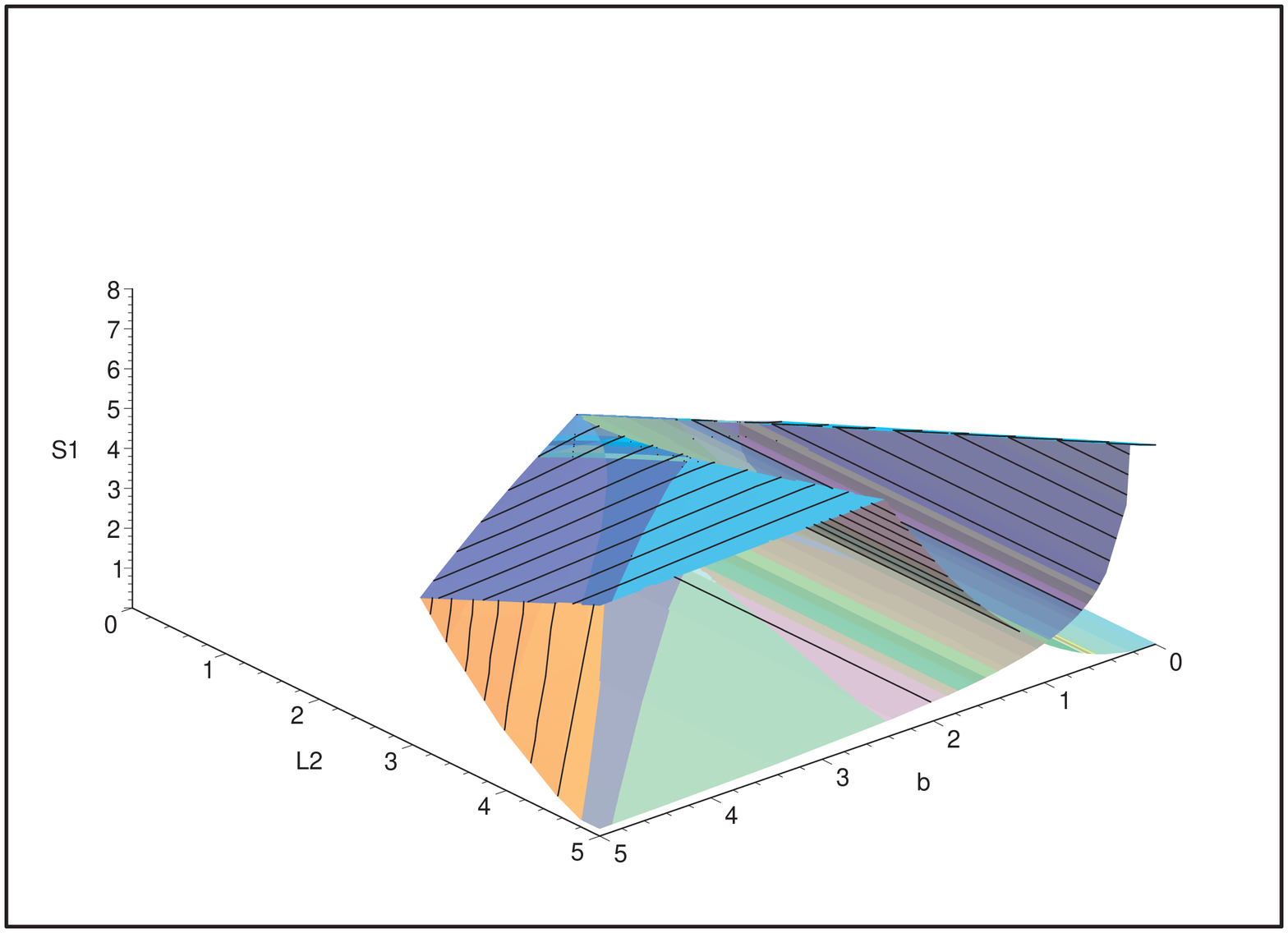}
\hspace{0.1\textwidth}
\includegraphics[height=0.4\textwidth,angle=-90]{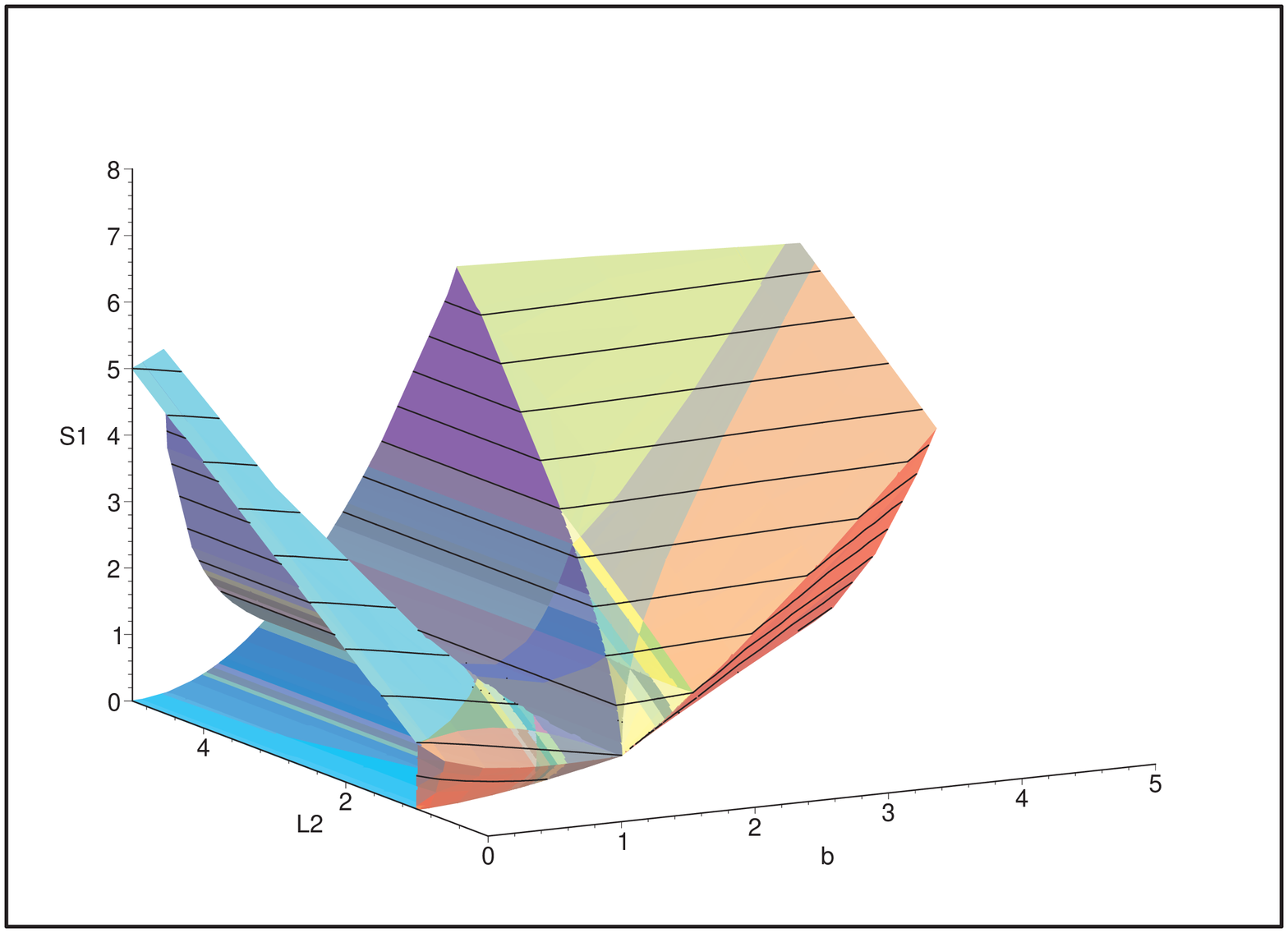}
\caption{The cells of $\mathbb{R}^3$ where the manipulator admits cuspidal configurations, front view (left) and back view (right).}
\label{figure:cusp}
\end{center}
\vspace*{-1cm}
\end{figure}

For our example, the process described in \ref{section:CAD} returns $344$ cells of dimension $3$ ( $\mathcal{U}_{3,1},...,\mathcal{U}_{3,344}$ ). We solve the system $\mathcal{S}$ for each of the 344 associated sample points, and we get always either $0$ or $16$ solutions. By selecting only the cells where the manipulator has $16$ cuspidal configurations, we obtain the 58 cells shown in Figure \ref{figure:cusp}.  Table \ref{table:bounds} provides the different formula bounding the three dimensional cells $\mathcal{U}_{3,1},...,\mathcal{U}_{3,344}$ and Table \ref{table:cells} represents the $58$ cells of Figure \ref{figure:cusp}, where the manipulator has $16$ cuspidal configurations.
\begin{table}
\vspace*{-0.5cm}
\tiny
\begin{center}
\caption{Formula describing the boundaries of the cells in Table \ref{table:cells}.}
\begin{tabular}{|p{5.4cm}|p{0.9cm}@{}p{5.2cm}|}
\hline
$b_1= 0, b_2= {\rm Root}(8\,b^6-11\,b^{4}+6\,b^2-1,2)$ &
$L_{2_1}(b)=$&${\rm Root}\left(\left( b^2+1 \right) ^{3}L_2^6-3\left( b^2+3b+1 \right)  \left( b^2-3b+1 \right)\left( b-1 \right)^2 \left( b+1 \right)^2 L_2^4+ \right.$ \\
$b_3= {\rm Root}(4\,b^2+b^6-3\,b^{4}-1, 2),\, b_4= {\rm Root}(b^{8}+3\,b^6+3\,b^{4}+b^2-1,2)$ &
&$\left. 3 \left( b^2+1 \right)\left( b-1 \right)^4 \left( b+1 \right) ^4L_2^2- \left( b-1 \right)^6 \left( b+1 \right) ^6, 2\right)$  \\
$b_5= {\rm Root}(-2\,b^4+b^6+3\,b^2-1,2),\, b_6= 1 / \sqrt2 ,\, b_7= 1 ,\, b_8= \sqrt(2)$  &
$L_{2_2}(b)=$&$1-b^2,\, L_{2_3}(b)= 1 / \sqrt{1-b^2},\, L_{2_4}(b)=(1-b^2)/b,\, L_{2_5}(b)=\infty$   \\
$b_9= {\rm Root}(2\,b^2+b^6-3\,b^4-1,2),\, b_{10}= {\rm Root}(b^{8}-b^6-3\,b^4-3\,b^2-1, 2)$  &
$L_{2_6}(b)=$&$ b^2 / \sqrt{1-b^2},\, L_{2_7}(b)= 1 / \sqrt{b^2-1},\, L_{2_8}(b)= b^2 / \sqrt{b^2-1}$\\
$b_{11}= {\rm Root}(-4\,b^4+b^6+3\,b^2-1, 2)$  &
$L_{2_9}(b)=$&$ (b^2-1)/b,\, L_{2_{10}}(b)=b^2-1$ \\
$b_{12}= {\rm Root}(b^6-6\,b^4+11\,b^2-8, 2),\, b_{13}= \infty$ &
& \\
\hline
\multicolumn{3}{|p {12cm}|}{$\rho_{1_1}(b,L_2)= {\rm Root}(-\rho_1^6b^6+3\,b^4\left(L_2^2b^2+1-L_2^2\right)\rho_1^4-3\,b^2\left(-7\,L_2^2b^2+7\,L_2^2+L_2^4+L_2^4b^4-2\,L_2^4b^2+1 \right) \rho_1^2 + \left( L_2^2b^2+1-L_2^2 \right) ^{3}, 2)$}\\
\multicolumn{3}{|p {12cm}|}{$\rho_{1_2}(b,L_2)= {\rm Root}(\rho_1^6+ \left( -3\,b^4+3\,L_2^2b^2-3\,L_2^2 \right)\rho_1^4 +\left( 21\,L_2^2b^6+3\,L_2^4b^4-6\,L_2^4b^2+3\,b^{8}-21\,L_2^2b^4+3\,L_2^4 \right)\rho_1^2+ \left( L_2^2b^2-b^4-L_2^2 \right)^{3}, 2)$} \\
\multicolumn{3}{|p {11.4cm}|}{$\rho_{1_2}(b,L_2)= b^2,\, \rho_{1_3}(b,L_2)= 1/b$} \\\hline
\end{tabular}
\label{table:bounds}
\end{center}
\vspace*{-0.5cm}
\end{table}

\begin{table}
\caption{Cells of $\mathbb{R}^3$ where the manipulator has cuspidal configurations.}
{\tiny
\noindent
\begin{tabular}{|p{1cm}||p {10 cm}|} \hline
$]b_1\,b_2[$&$   
                                                                                         (]L_{2_1}\, L_{2_2}[, ]\rho_{1_1}\,  \rho_{1_2}[),
                       (]L_{2_2}\, L_{2_3}[, ]\rho_{1_3}\,  \rho_{1_2}[),
                       (]L_{2_3}\, L_{2_4}[, ]\rho_{1_3}\,  \rho_{1_2}[),
                       (]L_{2_4}\, L_{2_5}[, ]\rho_{1_3}\,  \rho_{1_4}[)$\\[1ex]\hline
$]b_2\,b_3[$&$
                       (]L_{2_1}\, L_{2_6}[, ]\rho_{1_1}\,  \rho_{1_2}[),
                       (]L_{2_6}\, L_{2_2}[, ]\rho_{1_1}\,  \rho_{1_2}[),
                       (]L_{2_2}\, L_{2_3}[, ]\rho_{1_3}\,  \rho_{1_2}[),
                       (]L_{2_3}\, L_{2_4}[, ]\rho_{1_3}\,  \rho_{1_2}[),
                       (]L_{2_4}\, L_{2_5}[, ]\rho_{1_3}\,  \rho_{1_4}[)$\\[1ex]\hline
$]b_3\,b_4[$&$
                       (]L_{2_1}\, L_{2_6}[, ]\rho_{1_1}\,  \rho_{1_2}[),
                       (]L_{2_6}\, L_{2_2}[, ]\rho_{1_1}\,  \rho_{1_2}[),
                       (]L_{2_2}\, L_{2_4}[, ]\rho_{1_3}\,  \rho_{1_2}[),
                       (]L_{2_4}\, L_{2_3}[, ]\rho_{1_3}\,  \rho_{1_4}[),
                       (]L_{2_3}\, L_{2_5}[, ]\rho_{1_3}\,  \rho_{1_4}[)$\\[1ex]\hline
$]b_4\,b_5[$&$
                       (]L_{2_1}\, L_{2_6}[, ]\rho_{1_1}\,  \rho_{1_2}[),
                       (]L_{2_6}\, L_{2_2}[, ]\rho_{1_1}\,  \rho_{1_2}[),
                       (]L_{2_2}\, L_{2_4}[, ]\rho_{1_3}\,  \rho_{1_2}[),
                       (]L_{2_4}\, L_{2_3}[, ]\rho_{1_3}\,  \rho_{1_4}[),
                             (]L_{2_3}\, L_{2_5}[, ]\rho_{1_3}\,  \rho_{1_4}[)$\\[1ex]\hline
$]b_5\,b_6[$&$
                       (]L_{2_1}\, L_{2_2}[, ]\rho_{1_1}\,  \rho_{1_2}[),
                       (]L_{2_2}\, L_{2_6}[, ]\rho_{1_3}\,  \rho_{1_2}[),
                       (]L_{2_6}\, L_{2_4}[, ]\rho_{1_3}\,  \rho_{1_2}[),
                                                                                         (]L_{2_4}\, L_{2_3}[, ]\rho_{1_3}\,  \rho_{1_4}[),
                             (]L_{2_3}\, L_{2_5}[, ]\rho_{1_3}\,  \rho_{1_4}[)$\\[1ex]\hline
$]b_6\,b_7[$&$
                       (]L_{2_1}\, L_{2_2}[, ]\rho_{1_1}\,  \rho_{1_2}[),
                       (]L_{2_2}\, L_{2_4}[, ]\rho_{1_3}\,  \rho_{1_2}[),
                       (]L_{2_4}\, L_{2_6}[, ]\rho_{1_3}\,  \rho_{1_4}[),
                       (]L_{2_6}\, L_{2_3}[, ]\rho_{1_3}\,  \rho_{1_4}[),
                             (]L_{2_3}\, L_{2_5}[, ]\rho_{1_3}\,  \rho_{1_4}[)$\\[1ex]\hline
$]b_7\,b_8[$&$
                       (]L_{2_1}\, L_{2_9}[, ]\rho_{1_2}\,  \rho_{1_1}[),
                             (]L_{2_9}\, L_{2_{10}}[, ]\rho_{1_4}\,  \rho_{1_1}[),
                       (]L_{2_{10}}\, L_{2_7}[, ]\rho_{1_4}\,  \rho_{1_3}[),
                       (]L_{2_7}\, L_{2_8}[, ]\rho_{1_4}\,  \rho_{1_3}[),
                       (]L_{2_8}\, L_{2_5}[, ]\rho_{1_4}\,  \rho_{1_3}[)$\\[1ex]\hline
$]b_8\,b_9[$&$
                       (]L_{2_1}\, L_{2_9}[, ]\rho_{1_2}\,  \rho_{1_1}[),
                       (]L_{2_9}\, L_{2_7}[, ]\rho_{1_4}\,  \rho_{1_1}[),
                       (]L_{2_7}\, L_{2_{10}}[, ]\rho_{1_4}\,  \rho_{1_1}[),
                       (]L_{2_{10}}\, L_{2_8}[, ]\rho_{1_4}\,  \rho_{1_3}[),
                       (]L_{2_8}\, L_{2_5}[, ]\rho_{1_4}\,  \rho_{1_3}[)$\\[1ex]\hline
$]b_9\,b_{10}[$&$
                       (]L_{2_1}\, L_{2_7}[, ]\rho_{1_2}\,  \rho_{1_1}[),
                       (]L_{2_7}\, L_{2_9}[, ]\rho_{1_2}\,  \rho_{1_1}[),
                       (]L_{2_9}\, L_{2_{10}}[, ]\rho_{1_4}\,  \rho_{1_1}[),
                       (]L_{2_{10}}\, L_{2_8}[, ]\rho_{1_4}\,  \rho_{1_3}[),
                       (]L_{2_8}\, L_{2_5}[, ]\rho_{1_4}\,  \rho_{1_3}[)$\\[1ex]\hline
$]b_{10}\,b_{11}[$&$
                       (]L_{2_1}\, L_{2_7}[, ]\rho_{1_2}\,  \rho_{1_1}[),
                       (]L_{2_7}\, L_{2_9}[, ]\rho_{1_2}\,  \rho_{1_1}[),
                       (]L_{2_9}\, L_{2_{10}}[, ]\rho_{1_4}\,  \rho_{1_1}[),
                       (]L_{2_{10}}\, L_{2_8}[, ]\rho_{1_4}\,  \rho_{1_3}[),
                       (]L_{2_8}\, L_{2_5}[, ]\rho_{1_4}\,  \rho_{1_3}[)$ \\[1ex]\hline
$]b_{11}\,b_{12}[$&$
                       (]L_{2_1}\, L_{2_7}[, ]\rho_{1_2}\,  \rho_{1_1}[),
                       (]L_{2_7}\, L_{2_9}[, ]\rho_{1_2}\,  \rho_{1_1}[),
                       (]L_{2_9}\, L_{2_8}[, ]\rho_{1_4}\,  \rho_{1_1}[),
                       (]L_{2_8}\, L_{2_{10}}[, ]\rho_{1_4}\,  \rho_{1_1}[,
                       (]L_{2_{10}}\, L_{2_5}[, ]\rho_{1_4}\,  \rho_{1_3}[)$ \\[1ex]\hline
$]b_{12}\,b_{13}[$&$
                       (]L_{2_1}\, L_{2_9}[, ]\rho_{1_2}\,  \rho_{1_1}[),
                       (]L_{2_9}\, L_{2_8}[, ]\rho_{1_4}\,  \rho_{1_1}[),
                       (]L_{2_8}\, L_{2_{10}}[, ]\rho_{1_4}\,  \rho_{1_1}[),
                       (]L_{2_{10}}\, L_{2_5}[, ]\rho_{1_4}\,  \rho_{1_3}[)$ \\[1ex]\hline
\end{tabular} 
}
\label{table:cells}
\vspace*{-1cm}
\end{table}
\section{Conclusion}
We have proposed a general method to describe rigorously the design parameters for which a manipulator has cuspidal configurations. This method can be applied directly to other mechanisms, such as the ones studied in \cite{CR02,Wjmd98} for example. The tools used to perform the computations were implemented in a Maple library called Siropa\footnote{http://www.irccyn.ec-nantes.fr/~moroz/siropa/doc}. For 3D illustration purposes, we have detailed the main computations to be performed with manipulators satisfying $L_2 = L_3$. However, the proposed method allows one directly to solve the general case ($L_2\neq L_3$) by computing a discriminant variety of the system with $4$ parameters $b,L_2,L_3,\rho_1$, and by decomposing $\mathbb{R}^4$ with a CAD adapted to the discriminant variety. This description generalizes and completes the analyse done in \cite{HAPMieee09}. There is still some limitations though. In particular, when the system that defines the cuspidal configurations has no solution, it may mean that there exists a manipulator with no cuspidal configurations, but it may also mean that no manipulator can be assembled with these design parameters. Thus it is essential to be able to describe precisely the set of design parameter values for which  a manipulator can be assembled.
\begin{acknowledgement}
The research work reported here was made possible by SiRoPa ANR Project.
\end{acknowledgement}

\vspace{-1cm}

\renewcommand\section\sect

\end{document}